\documentclass[runningheads]{llncs}
\usepackage[T1]{fontenc}
\usepackage{graphicx,verbatim}
\usepackage{xcolor}
\usepackage{dsfont}
\usepackage{amsmath}
\usepackage{amssymb}
\usepackage{acronym}
\usepackage[capitalise]{cleveref}
\usepackage{siunitx}

\acrodef{IA}[IA]{Intracranial aneurysm}
\acrodef{AVM}[AVM]{arteriovenous malformation}
\acrodef{CFD}[CFD]{computational fluid dynamics}
\acrodef{MRI}[MRI]{magnetic resonance imaging}
\acrodef{MR}[MR]{magnetic resonance}
\acrodef{WSS}[WSS]{wall shear stress}
\acrodef{MRA}[MRA]{magnetic resonance angiography}
\acrodef{MLP}[MLP]{multilayer perceptron}
\acrodef{CNN}[CNN]{convolutional neural network}
\acrodef{GNN}[GNN]{graph neural network}
\acrodef{FNO}[FNO]{Fourier neural operator}
\acrodef{CNN}[CNN]{Convolutional Neural Network}
\acrodef{LSTM}[LSTM]{Long Short-Term Memory}
\acrodef{GRU}[GRU]{Gated Recurrent Unit}
\acrodef{NO}[NO]{Neural Operator}
\acrodef{FNO}[FNO]{Fourier Neural Operator}
\acrodef{DAFNO}[DAFNO]{Domain Agnostic Fourier Neural Operator}
\acrodef{FFT}[FFT]{Fast Fourier Transform}
\acrodef{iFFT}[iFFT]{inverse Fast Fourier Transform}
\acrodef{PDE}[PDE]{Partial Differential Equation}
\acrodef{PINN}[PINN]{Physics-Informed Neural Network}
\acrodef{EDSR}[EDSR]{Enhanced Deep Super-Resolution Network}
\acrodef{SRCNN}[SRCNN]{Super-Resolution Convolutional Neural Network}
\acrodef{SRNO}[SRNO]{Super-Resolution Neural Operator}
\acrodef{LoFNO}[LoFNO]{Localized Fourier Neural Operator}
\acrodef{PINN}[PINN]{Physics-Informed Neural Network}
\acrodef{DFI}[DFI]{Data-Driven Flow Interpolation}
\acrodef{RBF}[RBF]{Radial Basis Function}
\acrodef{PDE}[PED]{Partial Differential Equation}
\acrodef{LEP}[LEP]{Laplacian Eigevalues Prior}
\begin{document}
\title{Localized FNO for Spatiotemporal Hemodynamic Upsampling in Aneurysm MRI}
\titlerunning{Localized FNO for Spatiotemporal Upsampling}

\begin{comment}  %% Removed for anonymized MICCAI 2025 submission
\author{First Author\inst{1}\orcidID{0000-1111-2222-3333} \and
Second Author\inst{2,3}\orcidID{1111-2222-3333-4444} \and
Third Author\inst{3}\orcidID{2222--3333-4444-5555}}
%
\authorrunning{F. Author et al.}
% First names are abbreviated in the running head.
% If there are more than two authors, 'et al.' is used.
%
\institute{Princeton University, Princeton NJ 08544, USA \and
Springer Heidelberg, Tiergartenstr. 17, 69121 Heidelberg, Germany
\email{lncs@springer.com}\\
\url{http://www.springer.com/gp/computer-science/lncs} \and
ABC Institute, Rupert-Karls-University Heidelberg, Heidelberg, Germany\\
\email{\{abc,lncs\}@uni-heidelberg.de}}

\end{comment}

\author{
Kyriakos Flouris\textsuperscript{*}\inst{1}\orcidID{0000-0001-7952-1922} \and Moritz Halter\textsuperscript{*}\inst{1}\and
Yolanne Y. R. Lee\inst{2} \and Samuel Castonguay\inst{3}\inst{4} \and Luuk Jacobs\inst{5}  \and Pietro Dirix\inst{5} \and  Jonathan Nestmann\inst{5}\and Sebastian Kozerke\inst{5} \and
Ender Konukoglu\inst{1}}
\authorrunning{K. Flouris et al.}
\institute{
Biomedical Imaging Group, Computer Vision Lab, ETH Zurich
\\
\email{\{kflouris;mhalter,kender\}@ethz.ch}
\and
Department of Computer Science, University College London
\email{yolanne.lee.19@ucl.ac.uk}
\and
Institute of Environmental Engineering, Swiss Federal Institute of Technology, ETH Zurich
\and
Biodiversity and Conservation Biology Unit, Swiss Federal Institute for Forest, Snow and Landscape Research, WSL
\and
Institute for Biomedical Engineering, University and ETH Zurich
}
\renewcommand{\thefootnote}{\fnsymbol{footnote}}
\footnotetext[1]{These authors contributed equally to this work.}
\maketitle              % typeset the header of the contribution
\begin{abstract}

Hemodynamic analysis is essential for predicting aneurysm rupture and guiding treatment. While magnetic resonance flow imaging enables time-resolved volumetric blood velocity measurements, its low spatiotemporal resolution and signal-to-noise ratio limit its diagnostic utility. To address this, we propose the Localized Fourier Neural Operator (LoFNO), a novel 3D architecture that enhances both spatial and temporal resolution with the ability to  predict wall shear stress (WSS) directly from clinical imaging data. LoFNO integrates Laplacian eigenvectors as geometric priors for improved structural awareness on irregular, unseen geometries and employs an Enhanced Deep Super-Resolution Network (EDSR) layer for robust upsampling. By combining geometric priors with neural operator frameworks, LoFNO de-noises and spatiotemporally upsamples flow data, achieving superior velocity and WSS predictions compared to interpolation and alternative deep learning methods, enabling more precise cerebrovascular diagnostics.
The code, ablations and hyperparameters are available at: \url{https://github.com/moritz-halter/deepflow}.

% Authors must provide keywords and are not allowed to remove this Keyword section.

\end{abstract}

\section{Introduction}

\acp{IA} and \acp{AVM} are major causes of hemorrhagic strokes, leading to significant morbidity and mortality~\cite{van2007subarachnoid}. Timely surgical or endovascular interventions are critical to reducing rupture risk and preventing cerebral hemorrhage in high-risk individuals~\cite{neurolint16010005,pan2021bavnreview}. Accurate diagnostic imaging is essential for identifying at-risk patients, assessing disease severity, and guiding treatment planning~\cite{maupu2022imaging,hussein2020imaging}.

Current clinical imaging primarily captures morphological data, often overlooking hemodynamic parameters that could enable earlier rupture prediction~\cite{HAN202115,NICO202368}. Hemodynamic analyses based solely on morphology rely on costly \ac{CFD} simulations or simplified models derived from limited imaging slices, which may not fully capture the complexity of cerebral blood flow~\cite{maramkandam2024review}. Recent advances in 4D flow \ac{MR} imaging now allow time-resolved volumetric measurements of blood velocity vector fields throughout the brain~\cite{morgan20214d}, providing real data that was previously only available through \ac{CFD} simulations. These measurements enable detailed flow pattern analyses, collateral flow activation, arterial and venous pulsatility, pressure gradients, and \ac{WSS}, aiding in the detection and diagnosis of \ac{IA}~\cite{anu-rupture}. While 4D flow \ac{MR} imaging eliminates the computational burden of \ac{CFD} and the labor-intensive construction of geometric models, its low spatiotemporal resolution and signal-to-noise ratio remain major limitations. Interpolation methods can improve resolution, but they require manual selection of methods and parameters for each case, limiting automation and scalability.

% Neural networks offer the possibility for upsampling such images: a prominent example is the \ac{EDSR}~\cite{lim2017enhanced}, which are designed for single-image super-resolution tasks. The anatomical geometry of \ac{IA} can vary significantly between patients, making upsampling and therefore diagnosis and rupture prevention~\cite{anu-rupture} particularly challenging tasks. In addition, various efforts have been explored to directly predict hemodynamic parameters from imaging data, bypassing the computational bottlenecks of traditional \ac{CFD}. Early studies showed neural networks' potential for estimating \ac{WSS}, with \acp{CNN} enabling real-time predictions in coronary geometries and segmenting high shear stress regions~\cite{cnnaneurysms}.
% Notably, recent efforts have introduced \acp{GNN} that use surface meshes from \ac{CFD} simulations to estimate \ac{WSS}, achieving high accuracy on synthetic vascular geometries~\cite{Dupuy_Odier_Lapeyre_Papadogiannis_2023}.

Neural networks are widely used for medical image upsampling, with models like \ac{EDSR}\cite{lim2017enhanced} excelling in single-image super-resolution. However, the anatomical variability of \ac{IA} complicates both upsampling and clinical tasks like diagnosis and rupture prevention\cite{anu-rupture}. Beyond image enhancement, neural networks have been explored for predicting hemodynamic parameters directly from imaging data, bypassing the computational bottlenecks of \ac{CFD}.
%Early studies used \acp{CNN} for real-time \ac{WSS} estimation and high shear stress region segmentation~\cite{cnnaneurysms}, while more recent approaches applied \acp{GNN} to surface meshes from \ac{CFD} simulations, achieving high accuracy on synthetic vascular geometries~\cite{Dupuy_Odier_Lapeyre_Papadogiannis_2023}.
However, solving temporal dynamics and medical imaging super-resolution remains an open challenge.

% We propose a novel, domain-agnostic \ac{FNO} architecture~\cite{liu2024domain} that incorporates embeddings of the eigenvectors of the Laplacian operator on graphs, to which we refer as \ac{LoFNO}. Our method is designed to both upsample the spatial resolution and predict hemodynamic parameters directly from flow and \ac{MR} imaging data routinely acquired in clinical practice. While our method does not offer the wealth of information and simulation capabilities of \acp{CFD}, it offers an efficient and automated alternative that can leverage 4D flow \ac{MR} data.

% In this work, we offer three key contributions. First, \ac{LoFNO} integrates \textit{Laplacian eigenvectors} as geometric priors to enhance structural awareness, enabling the model to handle irregular, unseen geometries with greater fidelity. Second, it incorporates an \textit{\ac{EDSR}-based super-resolution layer} within the architecture, improving super-resolution capabilities and robustness to noise.
% This combination of geometric priors, advanced super-resolution techniques, and neural operator frameworks establishes a versatile and accurate approach for predicting hemodynamic parameters on complex geometries. Third, we present a practical implementation of \ac{LoFNO} on a synthetic 4D flow \ac{IA} dataset. \cref{fig:data_pipeline} provides an overview of the pipeline, including aneurysm extraction, \ac{CFD} data preparation for model training, and velocity and (\ac{WSS}) prediction.

We propose a novel, domain-agnostic \ac{FNO} architecture, \ac{LoFNO}, which incorporates embeddings of Laplacian eigenvectors to enhance geometric awareness~\cite{liu2024domain}. Designed to both upsample spatial resolution and temporal dynamics, enabling the prediction of intermediate time steps from sparse inputs. \ac{LoFNO} predicts hemodynamic parameters directly from routinely acquired flow from \ac{MR} imaging data, providing an efficient alternative to traditional \acp{CFD}. 

\begin{figure}[ht!]
    \centering
    \includegraphics[width=0.9\textwidth]{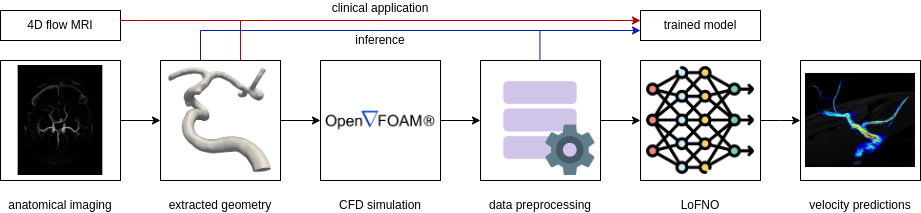}
    \caption{Method pipeline: Geometry and boundary conditions are extracted from cerebral angiograms. Simulations are run, the relevant section is segmented, Laplacian eigenvectors are computed, and noise is added. Flow and eigenvectors are sampled on a voxel grid for preprocessing. The model is trained to predict noise-free flow on a high-resolution grid and tested on unseen data. In clinical use, it can be applied directly to 4D flow MRI \textit{without} \ac{CFD} simulations.}
\label{fig:data_pipeline}
\end{figure}

Our approach introduces three key contributions: \textit{(1)} integrating \textit{Laplacian eigenvectors} as geometric priors to improve generalization across irregular vascular geometries,  \textit{(2)} incorporating an \textit{\ac{EDSR}-based super-resolution layer} to enhance image quality and robustness to noise, and  \textit{(3)} implementing \ac{LoFNO} on a synthetic 4D flow \ac{IA} dataset. Real data analysis is not feasible due to the need for high-quality training flow and geometry data, as obtaining very high-resolution 4D flow MRI is highly impractical. However, once trained, the model can be used to upsample low-resolution flow data, making it more applicable in a clinical setting. \cref{fig:data_pipeline} outlines the complete pipeline, from aneurysm extraction and \ac{CFD} data preparation to flow parameters prediction.
%Additionally, the method leverages a \textit{Cartesian voxel grid}, eliminating the need for mesh generation. 
% \begin{figure}
%     \centering
%     \includegraphics[width=0.9\textwidth]{images/pipeline.png}
%     \caption{Schematic of the method pipeline. \textit{I-II}: The geometry and boundary conditions are acquired from cerebral angiograms and reconstructed as part of the Aneurisk65 Dataset~\cite{sangalli2014aneurisk65}. \textit{III-IV}: From the simulated dataset, we manually segment the relevant section, calculate the Laplacian eigenvectors and add noise. \textit{V}: Surface normals are extracted. \textit{VI-VIII}: The flow, $\textit{VII},$ and eigenvectors, $\textit{VIII}$, are sampled on a Cartesian voxel grid. \textit{IX}: These inputs, along with added noise, are fed into our model. \textit{X}: The model is trained to predict the noise-free flow on a high-resolution grid. Using the model prediction and the surface normals, we calculate the \ac{WSS}, \textit{XI}.}

% \label{fig:data_pipeline}
% \end{figure}

\vspace{1em}

\noindent \textbf{Relevant Works} Deep neural networks are widely used for upsampling and super-resolution~\cite{cmf,srgan,lsd-ebm}, with \ac{SRCNN}~\cite{dong2015image} and \ac{EDSR}~\cite{lim2017enhanced} playing key roles.
 %Beyond image processing, deep learning has been applied to physics-constrained data, particularly in fluid dynamics~\cite{morimoto2021convolutional}.
 Physics-informed methods integrate physical priors into training, including \ac{PINN}s~\cite{cai2021physics} and other specialized frameworks~\cite{Liu_2024,wang2020physicsinformed,flouris2025}. While effective, these methods struggle with fixed discretization when solving continuous equations and face challenges in enforcing boundary conditions on complex geometries.

% Neural operators (\acp{NO})\cite{li2020neural} enable discretization-invariant learning by mapping between function spaces. Among them, Fourier Neural Operators (\acp{FNO})\cite{li2021fourier} leverage the \ac{FFT} for efficient \ac{PDE} solving, with various extensions improving performance and adaptability, including U-FNO~\cite{WEN2022104180}, Gabor-Filtered FNO~\cite{QI2024106239}, and $\varphi$-FEM-FNO~\cite{duprez2024varphi} but 
% they are constrained to fixed boundaries. \ac{DAFNO}\cite{liu2024domain} and Geo-\ac{FNO}~\cite{JMLR:v24:23-0064} address this limitation by enabling application to general geometries, while Incremental \ac{FNO}~\cite{zhao2022incremental} introduces adaptive training schedules to improve efficiency. Despite these advances, generalization to complex, unseen geometries remains a challenge, as these methods were not originally designed for intricate biological boundaries like those found in aneurysms.

\ac{DAFNO}\cite{liu2024domain} and Geo-\ac{FNO}\cite{JMLR:v24:23-0064} aim to alleviate the constraint of operator methods on fixed geometries by enabling applications to general geometries, while Incremental \ac{FNO}~\cite{zhao2022incremental} introduces adaptive training schedules to improve efficiency. Despite these advances, generalization to unseen geometries remains challenging, as these methods were not originally designed for intricate biological boundaries such as those in aneurysms. 

There are numerous  techniques  that leverage spectral coordinates for geometric processing~\cite{eigen1}~\cite{eigen2}, based on geometric deep learning~\cite{Bronstein2021GeometricDL}.  In this work, we introduce a novel architecture that builds upon \ac{DAFNO}, leveraging spectral embeddings~\cite{Bronstein2021GeometricDL}~\cite{eigen1} and incorporating \ac{EDSR} to efficiently upsample the spatial and temporal resoltuion of 4D flow imaging.

\section{Method}

The \ac{LoFNO} consists of an \ac{EDSR} module that takes low-resolution flow data and Laplacian eigenvectors as input, refining spatiotemporal resolution while integrating geometric features from spectral coordinates. The output, is then processed by a \ac{DAFNO} layer with implicit \ac{FNO}s, where a domain characteristic function ensures computations remain localized to the aneurysm geometry.

Operator methods enable learning generalized physical models across diverse domains. We aim to learn a mapping $\tilde{G}: \mathcal{A} \to \mathcal{U}$, where $\mathcal{A}(\mathbb{R}^{d_{in}})$ represents input velocity fields $u_i(x)$ and $\mathcal{U}(\mathbb{R}^{d_{out}})$ represents high-resolution output velocity fields $\hat{u}_i(x)$ over a patient-specific aneurysm domain $\Omega_i \subset \mathbb{R}^3$. This mapping is parameterized by $\theta$ and trained to approximate $\tilde{G}[u_i; \theta](x) \approx \hat{u}_i(x), \forall x \in \Omega_i$. Unlike traditional neural networks, the \ac{NO} maintains discretization invariance, allowing evaluation at arbitrary resolutions.

The \ac{FNO}~\cite{li2021fourier} leverages Fourier transforms for efficient modeling of spatial dependencies in partial differential equations. It applies a convolution-based layer where a tensor kernel $\kappa \in \mathbb{R}^{d_h \times d_h}$ with parameters $v^l$ operates over the computational domain $\bar{\Omega}$, which contains the aneurysm geometry $\Omega$. By utilizing the Fourier transform $\mathcal{F}$ and its inverse $\mathcal{F}^{-1}$, \ac{FNO} exploits the efficiency of \ac{FFT}/\ac{iFFT}. However, it requires inputs on a regular grid, making it unsuitable for irregular aneurysm geometries. To address this, a periodic domain grid $\mathds{O}$ is overlaid onto $\bar{\Omega}$, enabling Fourier-based computations but it has limited effectiveness.

\vspace{1em}

\noindent \textbf{Localizing to the Relevant Domains and Geomteric Prior}  
Mapping the aneurysm geometry $\Omega$ to the grid $\mathds{O}$ introduces out-of-domain voxels in $\mathds{O} \setminus \Omega$, increasing complexity and computational inefficiency. These out-of-domain voxels are trivial and  detract from the solution, particularly as boundary nodes are crucial for calculating \ac{WSS}. \ac{DAFNO}s~\cite{liu2024domain} extend \ac{FNO} to improve generalization across unseen domains while reducing computational overhead from $\mathds{O} \setminus \Omega$.

To enforce domain localization, $\chi(x)$ is defined as $\chi(x) = 1$ for $x \in \Omega$ and $\chi(x) = 0$ for $x \in \mathds{O} \setminus \Omega$. The Fourier layer is modified to ensure interactions remain within the physical domain, $\Omega$, while preserving \ac{FFT} efficiency. The operator is expressed compactly as:
\begin{align}
    \mathcal{J}^l[h](x) = \sigma\Big(
         \chi(x) \left( I(\chi(\cdot) h(\cdot); v^l) - h(x) I(\chi(\cdot); v^l) \right) \nonumber
         + W^l h(x) + c^l
    \Big),
\end{align}
where $I(\cdot; v^l) = \mathcal{F}^{-1} \left[ \mathcal{F}[\kappa(\cdot; v^l)] \cdot \mathcal{F}[\cdot] \right]$ represents the Fourier-based integral computation. $W^l \in \mathbb{R}^{d_h \times d_h}$ and $c^l \in \mathbb{R}^{d_h}$ are learnable parameters. This ensures domain-specific relevance while leveraging \ac{FFT} for efficient evaluation. Additionally, we adopt implicit \ac{FNO}s~\cite{you2022learning}, making learnable parameters layer-independent to mitigate overfitting and vanishing gradients.

To map the trained solution to complex and unseen geometries, we use \emph{spectral coordinates}. Laplacian eigenvectors are computed on a graph $\mathcal{G} = (\mathcal{V}, \mathcal{E})$, where $\mathcal{V}$ is the set of vertices and $\mathcal{E}$ the edges. The adjacency matrix $A \in \mathbb{R}^{N \times N}$ encodes connectivity, with $A_{ij} = 1$ if vertices $i$ and $j$ are connected, and the diagonal degree matrix $D$ has $D_{ii}$ as the vertex degree. The normalized graph Laplacian is defined as $L^{\text{sym}} = I - (D^+)^{1/2}A(D^+)^{1/2}$, where $I$ is the identity matrix and $D^+$ the Moore-Penrose pseudoinverse. Eigenvectors $\mathbf{v}_i$ and eigenvalues $\lambda_i$ are obtained from $L^{\text{sym}} \mathbf{v}_i = \lambda_i \mathbf{v}_i$, with eigenvalues sorted in descenting order. The first $k=32$ eigenvectors, corresponding to the largest nonzero eigenvalues, capture key geometric features and are normalized as $\mathbf{v}_i \leftarrow \mathbf{v}_i/|\mathbf{v}_i|_2, \forall i$.

 \vspace{1em}

\noindent \textbf{Enhanced architecture} Inspired from SRNO~\cite{superfno}, our approach first increases the resolution of the input data to prepare it for the Fourier layers. Instead of interpolation, we use an \ac{EDSR} network, designed for single-image super-resolution, leveraging convolutional layers and residual connections to handle noise in the input flow velocity data $u(x)$ sampled on a $d_x \times d_y \times d_z$ grid. This learnable upsampling improves resolution while reducing noise.

The \ac{EDSR} network takes as input the low-resolution flow velocity data concatenated with the selected eigenvectors $e(x)$ of the Laplacian operator. It outputs a high-resolution representation, which, along with the domain characteristic function $\chi(x)$, is fed into the \ac{DAFNO}. Within \ac{DAFNO}, the input is lifted by an \ac{MLP} layer $P$, then processed through Fourier layers, where \ac{FFT}, matrix multiplications with learnable parameters $v^l$, and nonlinear activations $\sigma$ are applied. The characteristic function $\chi(x)$ ensures domain-aware computations by restricting operations to relevant regions.

\begin{figure}
    \centering
\includegraphics[width=0.85\textwidth]{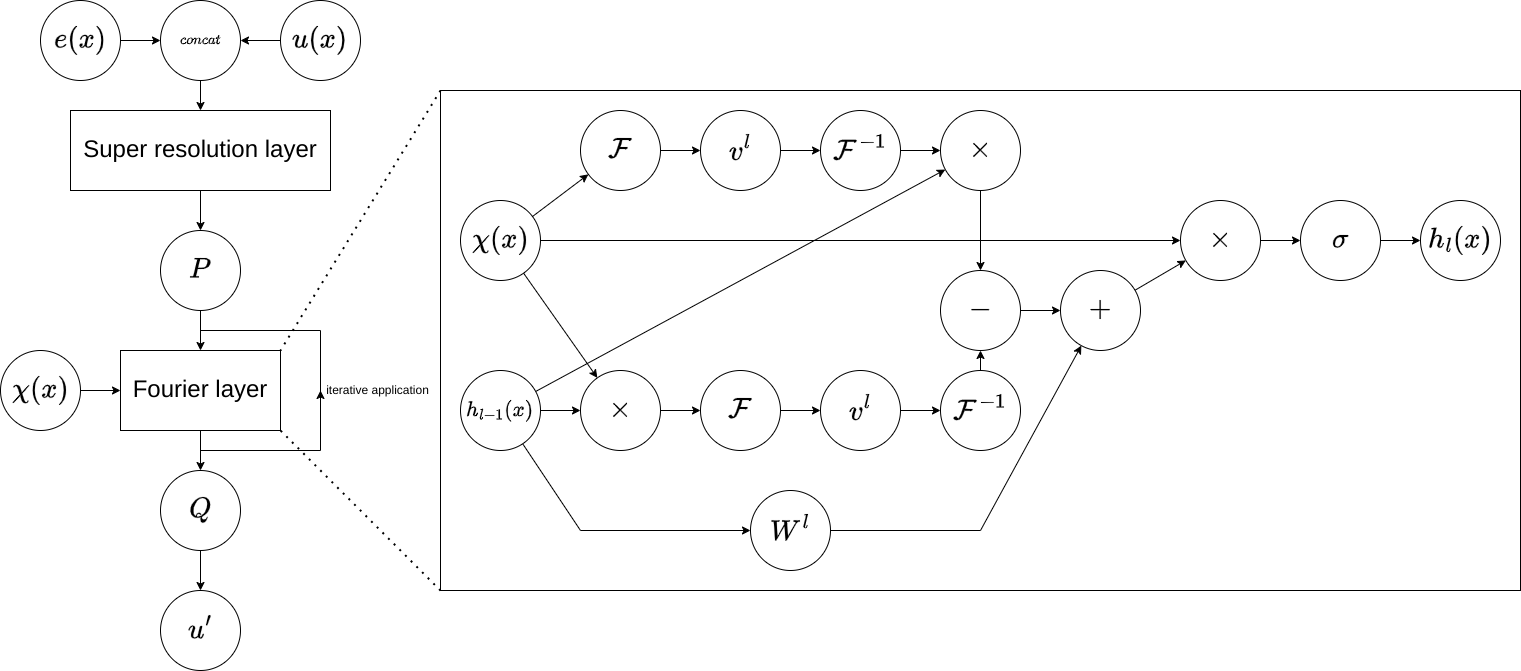}
    
    \caption{Computational graph for \ac{LoFNO}.
    % Both $e(x)$ a selection of eigenvectors and $u(x)$ the flow velocity with added noise are sampled on a regular grid with resolution $d_x \times d_y \times d_z$, while  $\hat{u}(x)$ the prediction for the noiseless flow velocity, the characteristic  $\chi(x)$ the domain characteristic function and $h_l(x)$ with $l \in \{0, \dots, L\}$ the fourier layer in and output are all sampled on a higher resolution grid with resolution $sd_x \times sd_y \times sd_z$. Lifting layer $P$ and projection layer $Q$ are \ac{MLP}s changing the dimensionality of their input respectively from $(3 + N_e)$ to $d_h$ and from $d_h$ to $3$ where $N_e$ the number of selected eigenvectors and $d_h$ the dimensionality in the fourier layers. Nodes $\mathcal{F}$ and $\mathcal{F}^{-1}$ performe the \ac{FFT} and \ac{iFFT}. Nodes $v^l$ performes a matrix multiplication with the learnable parameter $v^l \in \mathds{R}^{d_h \times d_h \times N_m \times N_m \times N_m}$ on the transformed input where the first $N_m$ modes are selected for each dimension. The node $W^l$ performs a point wise multiplication  with $W^l \in \mathds{R}^{d_h \times d_h}$. Nodes $+$, $-$, $\times$ and $\sigma$ perform point wise addition, subtraction, multiplication and non-linear activation respectively.   
    Both $e(x)$, a selection of eigenvectors, and $u(x)$, the flow velocity with added noise, are sampled on a regular grid,
    %with resolution $d_x \times d_y \times d_z$, 
    while $\hat{u}(x)$ (prediction for noiseless flow), $\chi(x)$ (domain characteristic function), and $h_l(x)$ (Fourier layer input/output, $l \in \{0, \dots, L\}$) are sampled on a higher resolution grid. 
    %$sd_x \times sd_y \times sd_z$.
    The lifting layer $P$ and projection layer $Q$ are \ac{MLP}s, mapping $(3 + N_e)$ to $d_h$ and $d_h$ to $3$, respectively, where $N_e$ is the number of eigenvectors and $d_h$ the Fourier layer dimensionality. Nodes $\mathcal{F}$ and $\mathcal{F}^{-1}$ perform \ac{FFT} and \ac{iFFT}, while $v^l$ applies matrix multiplication
    %with $v^l \in \mathds{R}^{d_h \times d_h \times N_m \times N_m \times N_m}$, 
    selecting the first $N_m$ modes for each dimension. The node $W^l$ performs pointwise multiplication with $W^l \in \mathds{R}^{d_h \times d_h}$, and $+$, $-$, $\times$, and $\sigma$ perform pointwise addition, subtraction, multiplication, and activation, respectively. The architecture and hyperparameters were optimized via ablation studies.}\label{fig:LoFNO}
\end{figure}

Finally, the output from the Fourier layers is projected to the target space through another \ac{MLP} layer $Q$, reducing dimensionality and producing the noiseless flow velocity prediction $\hat{u}(x)$. See \cref{fig:LoFNO} for a detailed schematic of the \ac{LoFNO} architecture.

\section{Experiments and Results}

%%% Brief CFD section %%%
 The Blood flow in the vessels was simulated by solving the three-dimensional, unsteady, incompressible Navier–Stokes equations. The blood was modeled as a Newtonian, incompressible fluid with a density of \(1060 \, \mathrm{kg/m^3}\) and a kinematic viscosity of \(3.5 \times 10^{-3} \, \mathrm{Pa \cdot s}\). In this study, the Navier–Stokes equations were solved using a large eddy simulation (LES) approach implemented in OpenFOAM v2212. The wall-adapting local eddy viscosity (WALE) model was chosen as the subgrid-scale model. Second-order central difference and backward Euler schemes were used for spatial and temporal discretization respectively. An adaptive time-stepping strategy was employed to optimize simulation efficiency. 
%The numerical simulations were performed on the Euler cluster operated by the High Performance Computing group at ETH Zürich.
%%%

Using these simulations, we have carried out two sets of experiments, the spatial super-resolution and temporal upsampling. We generated a dataset of 95 pulse flows imposed on geometries from the Aneurisk dataset~\cite{sangalli2014aneurisk65}. The dataset contains 3D reconstructions of internal carotid arteries and their associated aneurysms from cerebral angiographies. Simulations provided detailed hemodynamic parameters, including flow velocity, pressure, and \ac{WSS}, capturing the complex fluid dynamics within patient-specific vascular geometries. Testing was performed on an unseen subset of geometries and their respective eigenvectors.

\begin{figure} 
    \centering
    \includegraphics[width=0.99\textwidth]{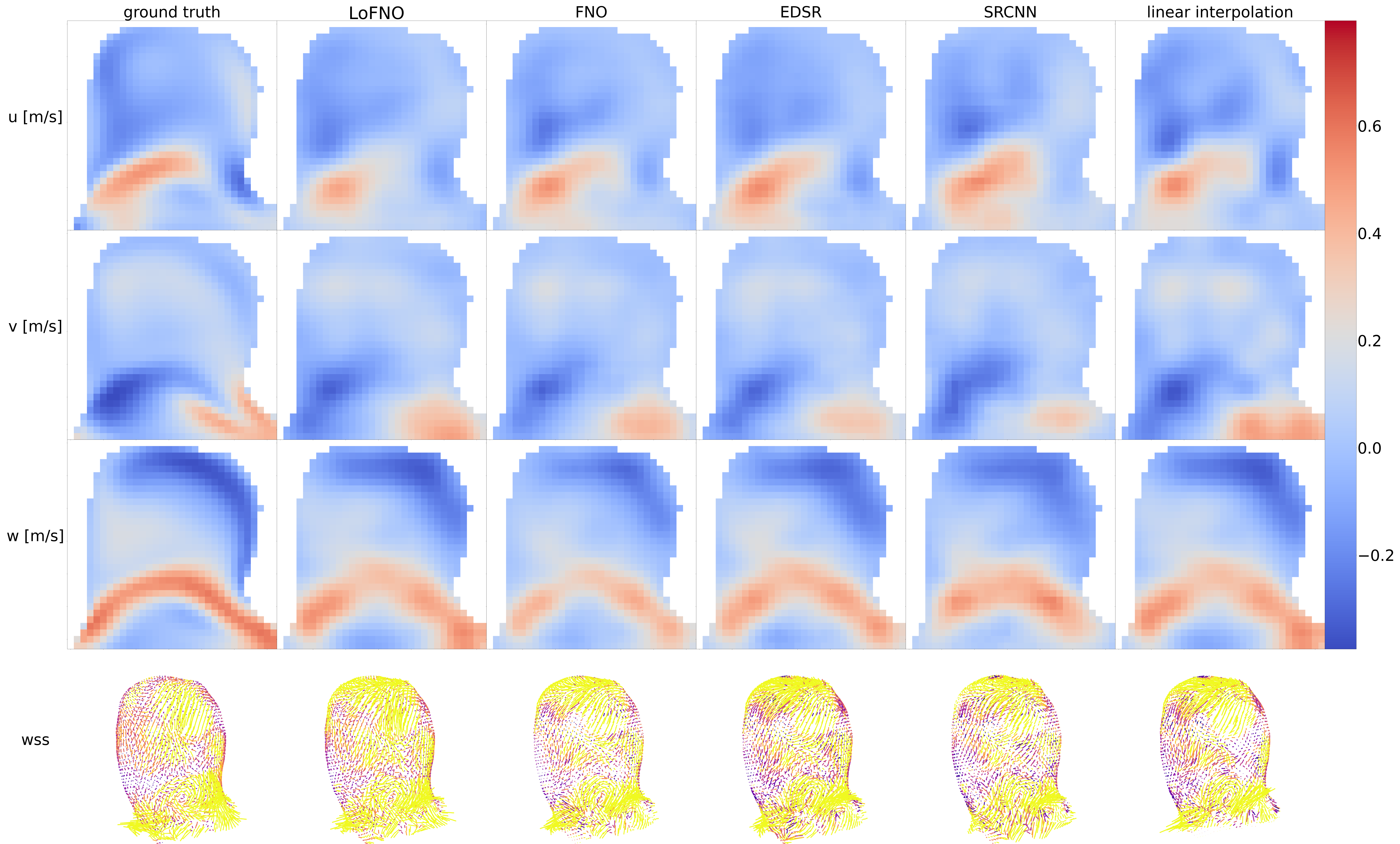}
    \caption{Time-snapshot of hemodynamics upsampling for Scale $\times4$: Top: Flow velocity components $u$, $v$, and $w$ in the $x$, $y$, and $z$ directions for  five models compared against the ground truth. Bottom: \ac{WSS} evaluated on the surface geometry, with the length of vectors colored black for the smallest to yellow for the largest.}
\label{fig:comparison}
\end{figure}

We selected 80 aneurysm-containing regions, computed Laplacian eigenvectors on the surface mesh, and resampled them onto a Cartesian grid using VTK's \texttt{vtkResampleWithDataSet}. Interior vessel points within subdomain $\Omega$ were marked, and white noise with a signal-to-noise ratio of 10 was added. Enforcing divergence-free interpolation was omitted as it caused artifacts, likely due to an overdetermined system from fixed points and the divergence-free constraint. Laplacian eigenvalues were computed on the VTK mesh, defining the graph, and resampled to the grid using point interpolation via \texttt{vtkPointInterpolator} with a Gaussian kernel to preserve boundary information.

 % The eigenvalues of the Laplacian were computed on the original VTK mesh, which defines the graph, and then resampled to the Cartesian grid using point-based interpolation as well, but with \texttt{vtkPointInterpolator} and a Gaussian Kernel to ensure enough points near the boundary contain information. 

% \begin{figure} 
%     \centeringcompute
%     \includegraphics[width=0.5\textwidth]{images/C0069_flow.png} 
%     \caption{Time-averaged blood flow ($u, v, w$) along the $x, y, z$ axes.} \label{fig:flow}
% \end{figure}

We trained several deep learning models, including \ac{SRCNN}, \ac{EDSR}, \ac{FNO}, and the proposed \ac{LoFNO}, conducting experiments. Additionally, we evaluated classical interpolation methods such as \ac{RBF} interpolation, \ac{DFI}, and linear interpolation. While \ac{PINN}s performed well with sparse data, they exhibited instability with denser inputs and proved inefficient for large 3D datasets, thus not included. 

Models were trained for 500 epochs, reaching near convergence. For spatial upsampling, models reconstructed a noiseless high-resolution flow field of either $32^3$ or $24^3$ from noisy low-resolution inputs of $16^3 $ or $8^3$, obtained by subsampling the high-resolution data across all $24$ timesteps.  For temporal upsampling, models reconstructed a full high-resolution $32^3 \times 24$ flow field from noisy, high-spatial-resolution but temporally undersampled inputs, including $32^3 \times 12$, $32^3 \times 6$, and $32^3 \times 1$, where for the latter the entire sequence was predicted from the initial timestep only. For all models, we minimized the relative $\text{loss} = \sum_{x\in\mathds{O}} \sum_t |u(x, t) - \hat{u}(x, t)| / |u(x, t)|$, where $u(x, t)$ is the ground truth flow velocity and $\hat{u}(x, t)$ the predicted velocity at spatial point $x$ and time $t$.

\vspace{1em}

\noindent \textbf{Results} For testing, we use a subset of 10 geometries that were unseen during training. The predicted results for the best-performing methods are shown in~\cref{fig:comparison}. While none of the methods achieved perfect reconstruction, as expected given the complexity of the problem and the unseen geometries, \ac{LoFNO} qualitatively demonstrated better performance than the other methods in reconstructing velocities and \ac{WSS}.

To evaluate model performance more concretely, we define the error function as $\text{err}f(x, t) = ||f(x, t) - \hat{f}(x, t)||_2$, which can be applied to flow velocity $u(x, t)$ or \ac{WSS}. Each model is benchmarked by averaging the error over all spatial and temporal dimensions across the test set. The differences are highlighted in the quantitative comparison in \cref{tab:super_res}, where complete \ac{LoFNO} outperformed other approaches, underscoring the significance of domain and geometric priors.

\begin{table}
    \centering
\caption{ Test-error, $L^{test}$, for hemodynamics parameters upsampling.}
\label{tab:super_res}
    \begin{tabular}{|l||l|l||l|l||l|l||l|}
        \hline
    Model & \multicolumn{2}{c|}{Scale $\times4$} & \multicolumn{2}{c|}{Scale $\times3$} & \multicolumn{2}{c|}{Scale $\times2$} & Training / \\ \cline{2-7}
    & $u$ & $wss$ & $u$ & $wss$ & $u$ & $wss$ & Evaluation\\ 
        \hline \hline
        Linear Interp. & 0.0684 & 1.0914 & 0.04661 & 0.5680 & 0.0497 & 0.8366 & $na / <1 \min$ \\
        RBF Interp. & 0.0813 & 1.4261 & 0.0563 & 0.6886 & 0.0591 & 0.9387 & $na / \sim 15 \si{\minute}$ \\
        DFI Interp. & 0.2054 & 2.8680 & 0.1339 & 1.3554 & 0.1864 & 2.5229 & $na / \sim 1 \si{\hour} $\\
        SRCNN & 0.0807 & 1.1081 & 0.0574 & 0.6527 & 0.0389 & 0.7969 & $\sim 15 \si{\minute} / \sim10 \si{\second}$ \\
        EDSR & 0.0683 & 1.1483 & 0.0460 & 0.5840 & 0.0300 & 0.6265 & $\sim 15 \si{\minute} / \sim 10 \si{\second}$ \\
        FNO\footnotemark[6] & 0.0647 & 0.9832 & 0.0388 & 0.4529 & 0.0272 & 0.5288 & $\sim 1 \si{\hour}/ \sim 10 \si{\second}$ \\
        \ac{LoFNO} $wo$ LEP  \footnotemark[7] & 0.0464 & 0.7699 & 0.0292 & 0.3750 & 0.0201 & 0.4220 & $\sim 1 \si{\hour}/ \sim 10 \si{\second}$ \\
        LoFNO  & \textbf{0.0452} & \textbf{0.7625} & \textbf{0.0291} & \textbf{0.3637} & \textbf{0.0198} & \textbf{0.4139} & $\sim 1 \si{\hour}/ \sim 10 \si{\second}$ \\
        \hline
    \end{tabular}
\end{table}

\footnotetext[6]{A \ac{FNO} with a preceding \ac{EDSR} layer}
\footnotetext[7]{\ac{LoFNO} without \ac{LEP}}

Combining neural operators with super-resolution networks yielded significant gains, as seen by the $\sim 25\%$ achieved by the \ac{LoFNO} $wo$ \ac{LEP}. The results are improved with the addition of the Laplacian eigenvectors as geometric priors. Noteworthily, not only does \ac{LoFNO} improve on the aggregated metrics, but the results consistently outperform the other methods for every individual test case. The localized geometry nature of our framework enhanced robustness and adaptability.
%Given the high noise levels in the input data, all machine learning approaches significantly outperformed classical interpolation methods.
While all methods train efficiently on modern hardware, interpolation techniques are slower during evaluation. We have also trained and tested on noiseless data, yielding similar trends, but omitted it due to limited real-world applicability.

\begin{table}
    \centering
    \caption{Test-error, $L^{test}$, for hemodynamics parameters temporal upsampling.}
    \label{tab:super-res-t}
    \begin{tabular}{|l||l|l||l|l||l|l||l|}
        \hline
        Model & \multicolumn{2}{c|}{Scale $\times2$} & \multicolumn{2}{c|}{Scale $\times4$} & \multicolumn{2}{c|}{Prediction} & Training / \\ \cline{2-7}
        & $u$ & $wss$ & $u$ & $wss$ & $u$ & $wss$ & Evaluation\\
        \hline \hline
        Linear Interp. & 0.0632 & 0.5631 & 0.0699 & 0.6107 & 0.1145 & 0.9520 & $na / <1 \min$ \\
        \ac{SRCNN} & 0.0461 & 0.7141 & 0.0607 & 0.8298 & 0.0600 & 0.8082 & $\sim 15 \si{\minute} / \sim 10 \si{\second}$ \\
        \ac{EDSR} & 0.0317 & 0.4689 & 0.0459 & 0.5765 & 0.0624 & 0.7671 & $\sim 15 \si{\minute} / \sim 10 \si{\second}$ \\
        \ac{FNO}\footnotemark[6] & 0.0384 & 0.4367 & 0.0382 & 0.4247 & 0.0678 & 0.7546 & $\sim 1 \si{\hour}/ \sim 10 \si{\second}$ \\
        \ac{LoFNO} $w/o$ \ac{LEP}\footnotemark[7] & 0.0245 & 0.2987 & 0.0315 & 0.3684 & 0.0531 & 0.6028 & $\sim 1 \si{\hour}/ \sim 10 \si{\second}$ \\
        \ac{LoFNO} & \textbf{0.0195} & \textbf{0.2723} & \textbf{0.0280} & \textbf{0.3424} & \textbf{0.0536} & \textbf{0.5941} & $\sim 1 \si{\hour}/ \sim 10 \si{\second}$ \\
        \hline
    \end{tabular}
\end{table}

For temporal super-resolution, our model showed even greater improvements, as seen in \cref{tab:super-res-t}. Even standalone \ac{FNO} performed well across all tested scales. However, despite full spatial resolution being available, augmenting \acp{FNO} with domain or geometric priors, such as the domain characteristic function and Laplacian eigenvectors i.e. the \ac{LoFNO}, further enhanced performance.

% A particularly challenging case is the complete prediction task i.e. \(32^3\times1\) to \(32^3\times24\), where 23 future time steps are predicted from a single initial one. Despite its complexity, all machine learning frameworks maintained reasonable accuracy, whereas interpolation methods suffered a severe performance drop. Even when given an extra $25^{th}$ time step, interpolation remained far less accurate than machine learning models.
% This disparity stems from the \ac{CFD} simulation applying the same pressure pulse at the vessel inlet %(\myfigref{fig:u-vs-t})
% . Machine learning models, trained on slight variations of these pulses, effectively learned the underlying dynamics, enabling accurate predictions even for unseen vessel geometries. In contrast, interpolation methods, lacking prior knowledge of the system, relied solely on given data points, limiting their predictive capability.

The complete prediction task, \(32^3\times1\) to \(32^3\times24\), where 23 future time steps are predicted from a single initial one, is particularly challenging. Despite this, all machine learning models, and especially our method, maintained reasonable accuracy, while interpolation methods suffered a severe performance drop. Even with an extra \(25^{th}\) time step, interpolation remained far less accurate. This disparity arises because the \ac{CFD} simulation applies the same pressure pulse at the vessel inlet. Machine learning models, trained on slight pulse variations, effectively captured the underlying dynamics, enabling accurate predictions even on unseen geometries. In contrast, interpolation methods, lacking system knowledge, relied only on given data points, limiting their predictive capability.

% \section{Limitations}
% This study has several limitations.  While the \ac{FNO} offers advantages in efficiency and scalability, it may be less flexible compared to generative methods for handling diverse data distributions. Additionally, our approach is currently constrained by the reliance on \ac{CFD} data for training, which may not fully capture the variability and complexity of real-world scenarios. Finally, while the method's use of interpolation on a Cartesian grid eliminates the need for mesh generation, it may restrict adaptability to irregular geometries, potentially limiting its broader applicability in certain clinical settings.
% \section {Limitations}
\vspace{1em}
%This study and method has several limitations. % it may be less flexible than generative methods in handling diverse data distributions. Moreover, our approach 
\noindent \textbf{Limitations} While \ac{LoFNO} offers efficiency and scalability, it depends on \ac{CFD} data for training, as no real data exist for this task, which may not capture real-world variability. These methods also lack inherent temporal mechanisms, limiting interpretability for dynamic flow modeling. We evaluated several super-resolution alternatives and found that the chosen methods, including the gold standard FNO, consistently matched or outperformed others. We were also constrained to methods that could be adapted to large 3D datasets.

\section{Conclusions}
We proposed \ac{LoFNO}, a novel architecture that %integrates Laplacian eigenvectors as geometric priors and an \ac{EDSR} layer for enhanced resolution. \ac{LoFNO} 
outperforms interpolation and alternative methods in test-error for spatial and temporal upsampling of velocity and \ac{WSS} predictions. 
Our focus in this work is to address a key methodological challenge: super-resolving fluid dynamics variables in complex boundaries remains unsolved, even with large-scale, high-fidelity simulated data. Generalizing to  unseen geometries is not only relevant to aneurysms but to many other problems. The proposed model directly targets this. To demonstrate its effectiveness, we focus on quantitative evaluation using high-resolution simulated data, where ground truth is available. This assessment would not have been possible with clinical data because the ground truth cannot be easily acquired.
% The trained model can be used to upsample 4D flow MRI. This approach has the potential to significantly improve clinical assessment and management of cerebrovascular pathologies like intracranial aneurysms and arteriovenous malformations, which present challenges due to their anatomical and hemodynamic complexity. 
By enabling precise, non-invasive prediction of hemodynamic parameters, \ac{LoFNO} can improve clinical assessment, enhance disease progression predictions, and optimize therapeutic outcomes, ultimately advancing patient care and safety.

%
% ---- Bibliography ----
%
% BibTeX users should specify bibliography style 'splncs04'.
% References will then be sorted and formatted in the correct style.
%
\bibliographystyle{splncs04}
% \bibliography{mybibliography}
%
 \bibliography{main}
% \begin{thebibliography}{8}
% \bibitem{ref_article1}
% Author, F.: Article title. Journal \textbf{2}(5), 99--110 (2016)

% \bibitem{ref_lncs1}
% Author, F., Author, S.: Title of a proceedings paper. In: Editor,
% F., Editor, S. (eds.) CONFERENCE 2016, LNCS, vol. 9999, pp. 1--13.
% Springer, Heidelberg (2016). \doi{10.10007/1234567890}

% \bibitem{ref_book1}
% Author, F., Author, S., Author, T.: Book title. 2nd edn. Publisher,
% Location (1999)

% \bibitem{ref_proc1}
% Author, A.-B.: Contribution title. In: 9th International Proceedings
% on Proceedings, pp. 1--2. Publisher, Location (2010)

% \bibitem{ref_url1}
% LNCS Homepage, \url{http://www.springer.com/lncs}, last accessed 2023/10/25
% \end{thebibliography}
\end{document}